\documentclass[]{acmart}
\AtBeginDocument{%
  \providecommand\BibTeX{{%
    \normalfont B\kern-0.5em{\scshape i\kern-0.25em b}\kern-0.8em\TeX}}}

\usepackage[utf8]{inputenc}
\usepackage{graphicx}
\usepackage{hyperref}

\usepackage{multirow}

\usepackage{tikz}
\usepackage{tikz-qtree}
\usetikzlibrary{automata, positioning, arrows}

\usepackage{gb4e}
\noautomath

\acmYear{Under review}
\acmJournal{TALLIP}
\acmMonth{0}

\begin{document}

\title{Towards Finite-State Morphology of Kurdish}

\author{Sina Ahmadi}
\email{sina.ahmadi@insight-centre.org}
\orcid{0000-0001-7904-6551}
\affiliation{%
  \institution{Insight Centre for Data Analytics, National University of Ireland Galway}
  \streetaddress{Galway Business Park}
  \city{Galway}
  \country{Ireland}
  \postcode{H91 AEX4}
}

\author{Hossein Hassani}
\orcid{0000-0002-8899-4016}
\affiliation{%
  \institution{University of Kurdistan Hewlêr}
  \streetaddress{30M Avenue}
  \city{Erbil}
  \country{Iraq}}
\email{hosseinh@ukh.edu.krd}

\renewcommand{\shortauthors}{Ahmadi and Hassani}

\begin{abstract}
Morphological analysis is the study of the formation and structure of words. It plays a crucial role in various tasks in Natural Language Processing (NLP) and Computational Linguistics (CL) such as machine translation and text and speech generation. Kurdish is a less-resourced multi-dialect Indo-European language with highly inflectional morphology. In this paper, as the first attempt of its kind, the morphology of the Kurdish language (Sorani dialect) is described from a computational point of view. We extract morphological rules which are transformed into finite state transducers for generating and analyzing words. The result of this research assists in conducting studies on language generation for Kurdish and enhances the Information Retrieval (IR) capacity for the language while leveraging the Kurdish NLP and CL into a more advanced computational level.  
\end{abstract}




\maketitle

\section{Introduction}
Morphology is the study of the structure of words. The morpheme is a phonemically defined segment of speech or set of segments of speech with a constant range of meaning. Morphological analysis is a central task in language processing that can take a word as input and detect the various morphological entities in the word and provide a morphological representation of it.

\begin{figure}
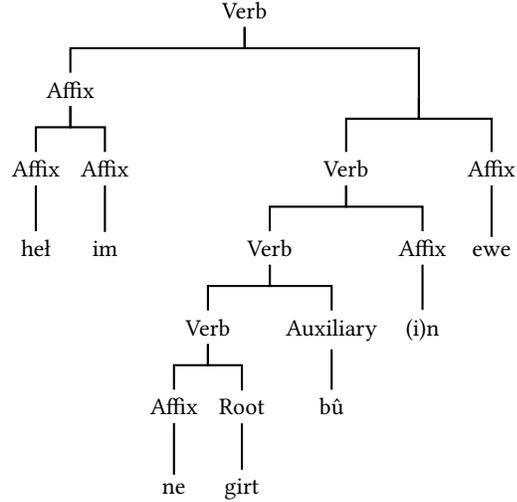

\centering
\tikzset{edge from parent/.style={thick,draw,edge from parent path={(\tikzparentnode.south)-- +(0,-8pt)-| (\tikzchildnode)}}}
\Tree [.Verb 
        [
            .Affix [.Affix heł ] [.Affix im ]
        ]
        [
            [.Verb 
                [.Verb [.Verb [.Affix ne ] [.Root girt ] ]
                    [
                    .Auxiliary bû
                    ]
                ]
                [.Affix (i)n ]
            ]
            .Affix [.Affix ewe ]
        ]
        ]
%\Tree [.Verb 
%        [.Affix heł ]
%        [.Verb [.Verb [.Verb [Root girt]  [Aux bû]] [.Affix (i)n] ] [.Affix ewe ]
%        ]
%    ]
\caption{Morphological tree of "hełimnegirtbûnewe"}
\label{morpho_tree}
\end{figure}

In the case of morphologically rich languages, multiple types and levels of information may be available at the word level. The lexical information for each word form may be augmented with information concerning the grammatical function of the word in the sentence, pronominal clitics, its grammatical relations to other words, inflectional affixes, and so on \cite{tsarfaty2013parsing}. In Kurdish, many of these notions are expressed by inflectional affixes. For instance, the stem of a transitive verb can be preceded or succeeded by affixes to indicate the subject, the direct object, tense and grammatical mood. Figure \ref{morpho_tree} illustrates the morphological tree of the verb "\textit{hełimnegirtbûnewe}" meaning "(I) had not retaken them". 

Analysing morphologically rich languages is a challenging task regardless of the analysis technique implemented. Our main objective in this paper is to demonstrate the capacity of Finite-State Transducers (FSTs) as a method towards computational morphology in the Kurdish language. We focus on Sorani Kurdish as one of the widely spoken and written dialects of Kurdish \cite{salavati2018building}. Given the complexity of the Kurdish morphology, we have only covered the major inflectional morphemes in the current study, namely nouns, verbs, adjectives, and adverbs. This project is also an attempt to address three items of the identified gaps in the Kurdish BLARK (Basic Language Resource Kit)~\cite{hassani2018blark}, namely morphological analysis, morphological synthesis, and text generation. 

In our approach to the morphological study of Kurdish, we follow the trend which is followed in the Natural Language Processing (NLP) realm. This approach has been described in the related literature, for example, by Jacquemin and Tzoukermann~\cite{jacquemin1999nlp}. That is, we are aware of the discussion and debates among the linguists upon the meaning of morphemes and lexemes and also the debate that is going on among the linguists in how to interpret  these concepts~\cite{nemo2003morphemes,beard2005lexeme}. However, we focus on the practical outcome which could help in the advancement of the resources and tools for Kurdish NLP. 

Finite-State Transducers (FSTs) are well-known apparatus in various sectors of computational field including NLP and CL \cite{karttunen2000applications}, in a variety of tasks such as machine translation~\cite{civera2005novel,forcada2011apertium} and computational morphology~\cite{altantawy2011fast}. In comparison to the traditional computational morphology based on lexicon, FSTs are proven to be more effective instruments concerning morphologically rich languages~\cite{altantawy2011fast}. One of the main obstacles that have hindered the progress in morphological analysis for Kurdish was lack of electronic lexicographic resources. Recently, Ahmadi et al. \cite{ahmadi2019lex} presented three electronic lexicographic resources for Kurdish covering the Hawrami, Sorani and Kurmanji dialects with enriched information such as part-of-speech. Therefore, as a preliminary study, we could construct a set of FST-based rules for four grammatical categories in Sorani Kurdish, namely nouns, adjectives, adverbs and verbs. For this purpose, We use Stuttgart Finite-State-Transducer (SFST)~\cite{schmid2005programming} as the experimental environment. 
%dataset for Verbs, Nouns, Adjective, and Adverbs of Sorani, 
% we extracted from the lexicographic dataset developed by Ahmadi et al.~\cite{ahmadi2019lex}.

The rest of this paper is organized as follows: Section \ref{relatedwork} reviews the literature and addresses the related work. Section \ref{Kurdishmorphology} presents the Kurdish morphology. In section \ref{implementation}, we describe the implementation of our morphological analyser for various morphological categories of Sorani Kurdish. And finally, the paper is concluded in Section \ref{conclusion} and a few idea for future research are proposed.

\section{Related Work}
\label{relatedwork}

Computational morphology using FSTs has been a topic of research since the 1980s \cite{karttunen2001short}, and a variety of tools have been developed for this purpose~\cite{beesleyFSM}. The Finite-State Morphology (FSM) has covered both morphological analysis and generation aspects of computational morphology addressing various languages some of which have been well-equipped by NLP and CL tools and some not. While resourceful languages such English~\cite{minnen2001applied,beesleyFSM} and German~\cite{schmid2005programming} are front-runners in FSM studies, we also observe similar scholarly attempts regarding other languages which might not be considered as widely-studied as English or German such as Uralic languages~\cite{novak2015model}, Arabic~\cite{soudi2007arabic} and Persian~\cite{megerdoomian2000persian,arabsorkhi-shamsfard-2006-unsupervised}. This is also correct for less-studied languages such as Croatian~\cite{mihajloviccomputational}. These examples denote the existence of wider attention on the area of computational morphology in NLP and CL whether by using FSTs or by following other approaches.

Despite its crucial role in the language analysis and generation, computational morphology has not received noticeable attention in Kurdish NLP and CL except for a few efforts~\cite{hassani2018blark}. Walther and Sagot~\cite{walther2010developing} and Walther et al.~\cite{walther2010fast} present a methodology and preliminary experiments on constructing a morphological lexicon for Sorani and Kurmanji in which a lemma and a morphosyntactic tag are associated with each known form of the word.  For this purpose, the Alexina framework \cite{sagot2010lefff} is used. 

Computational morphology of Kurdish has been partially addressed in related NLP and CL tasks. Hosseini et al.~\cite{hosseniKSLexicon} suggest a formulation for Sorani morphology used in creating a Sorani lexicon. Salavati et al.~\cite{salavati2018building} report challenges in Sorani Kurdish lemmatization and spelling error correction due to morphological complexity. \cite{gokirmak2017dependency} carries out a morphological analysis for creating a dependency treebank for Kurmanji Kurdish. 
 
%Regarding the other widely spoken Kurdish dialect, in their research about the regional variations in Kurmanji, Öpengin and Haig~\cite{haig2014regional} showed that morphologically there are significant differences among varieties for Kurmanji sometimes because of their regional border with Sorani speaking regions. They have given examples of morphological, grammatical, lexical and phonological differences among these varieties.
 
The aforementioned situation indicates that the lack of Kurdish computational morphology is an area that not only requires but also deserves a sizeable effort by and collaboration among interested scholars. 

% ******************************************************************************************
\section{Kurdish Morphology}
\label{Kurdishmorphology}

Kurdish is an Indo-European language which is spoken by approximately 30 million people in different countries~\cite{ahmadi2019lex}.
Haig and Matras~\cite{haig2002kurdish} provide a brief description of the structural properties of Kurdish. As a multi-dialect language, Kurdish dialects have different grammatical features and vocabulary sets~\cite{hassani2016automatic}. The differences in grammar and vocabulary vary among the dialects~\cite{jugel2014linguistic,haig2014introduction}. In several cases, the differences are significant while in the others, they are trivial~\cite{hassani2016automatic}. Equally important, the language is written in different scripts with no standard orthography~\cite{ahmadi2019wergor}. This enforces the computational morphology for Kurdish to be dialect-focused, at least in the first steps.

Kurdish is considered a morphologically rich language for which grammatical relations are indicated by changes in the word forms and modifying morpheme~\cite{littell2016named,zivingi2019comparative}. Unlike most Indo-European languages which are considered fusional languages, Kurdish is also characterized as a partially agglutative language \cite{khalid2015new}. Agglutination refers to a linguistic process in which various word forms are created by stringing morphemes together. 

Our focus in this paper is on four grammatical categories, namely verbs, nouns, adjectives, and adverbs. These categories are briefly described as follows.

\subsection{Nouns}

The absolute form of a noun in Kurdish is the form without any affixes which represents a generic meaning of the word \cite{thackston2006sorani}. This form is the lemma provided in the dictionary. The inflection of nouns is mostly carried out with suffixes to indicate number, definiteness, indefiniteness, demonstratives and gender. Unlike Kurmanji and Hawrami dialects, Sorani and Southern Kurdish dialects do not specify genders through morphological inflection. However, there are a few exceptions in some of the subdialects of the two latter dialects, such as the Mukryani subdialect of Sorani where nouns can have genders in specific cases. For instance, in the sentences "\textit{deçime małe Zeynebî}" and "\textit{le kin Ferhadê}", "\textit{Zeyneb}" and "\textit{Ferhad}" respectively as feminine and masculine proper names are inflected with the "î" and "ê" suffixes to represent the gender.

\begin{table}[h]
\centering
\begin{tabular}{l|l|l}
\hline
Noun form                      & Singular   & Plural       \\ \hline \hline
\multirow{2}{*}{absolute}      & \textit{naw}        & -            \\
                               & \textit{derga}      & -            \\\hline
\multirow{2}{*}{indefinite}    & \textit{naw\textbf{êk}}      & \textit{naw\textbf{an}}        \\
                               & \textit{derga\textbf{yek} }  & \textit{derga\textbf{yan}}     \\\hline
\multirow{2}{*}{definite}      & \textit{naw\textbf{eke}}     & \textit{naw\textbf{ekan}}      \\
                               & \textit{derga\textbf{ke} }   & \textit{ derga\textbf{kan}}     \\\hline
\multirow{2}{*}{demonstrative} & \textit{em naw\textbf{e} }   & \textit{em naw\textbf{ane}}    \\
                               & \textit{em derga\textbf{ye}} & \textit{em derga\textbf{yane}} \\\hline
\end{tabular}
\caption{Inflected forms of "\textit{naw}" (name) and "\textit{derga}" (door) in Sorani Kurdish}
\label{tab_nouns_inflection}
\end{table}

Table \ref{tab_nouns_inflection} describes the inflection of the two words "\textit{naw}" (name) and "\textit{derga}" (door) where the morphemes are highlighted in bold. As a result of the inflection, phonological changes, such as dropping a vowel or adding an auxiliary one, may happen when two vowels emerge consecutively. For instance, such a change can be observed in "\textit{dergaye}" where "\textit{y}" appears between the lemma "\textit{derga}" and the demonstrative suffix "\textit{e}".

\subsection{Adjectives and adverbs}

Adjectives in Sorani Kurdish are inflected according to the modified noun. Predicative adjectives in Kurdish are mostly inflected based on their syntactic role. In addition, the comparative and superlative degrees of an adjective are respectively made by suffixes "\textit{tir}" and "\textit{tirîn}". On the other hand, the attributive adjectives follow a richer morphological representation depending on the inflection of the noun. The main noun-adjective construction in Sorani Kurdish is carried out using the \textit{Izafa} morpheme. The Izafa is an "\textit{î}" or "\textit{y}" (following a vowel) appearing between the noun and the adjective. However, there are other types of making attributive adjectives which are differing in the noun-adjective construction and the placement of the definiteness or number suffixes \cite{thackston2006sorani}. Table \ref{tab_adjectives} provides an example of the attributive adjective "\textit{ciwan}" (beautiful) for the lemma "\textit{guł}" (flower) in various forms.

\begin{table}[h]
\begin{tabular}{l|l|l|l|l}
\hline
\multirow{3}{*}{Noun form} & \multicolumn{2}{c|}{Loose-Izafa}   & \multicolumn{2}{c}{Close-Izafa}   \\ \hline   
                           & Singular       & Plural           & Singular       & Plural           \\\hline\hline
absolute                   & \textit{guł\textbf{î} ciwan}     & -                & \textit{guł\textbf{e} ciwan}     & -                \\
indefinite                 & \textit{guł\textbf{êki} ciwan}   & \textit{guł\textbf{anî} ciwan}     & \textit{guł\textbf{e} ciwan\textbf{êk}}   & \textit{guł\textbf{e} ciwan\textbf{an}}     \\
definite                   & \textit{guł\textbf{eke} ciwan}   & \textit{guł\textbf{ekanî} ciwan}   & \textit{guł\textbf{e} ciwan\textbf{eke}}  & \textit{guł\textbf{e} ciwan\textbf{ekan}}   \\
demonstrative              & \textit{em guł\textbf{e} ciwan\textbf{e}} & \textit{em guł\textbf{e} ciwan\textbf{ane}} & \textit{em guł\textbf{e} ciwan\textbf{e}} & \textit{em guł\textbf{e} ciwan\textbf{ane}} \\\hline
\end{tabular}
\caption{Inflected forms of the adjective "\textit{ciwan}" (beautiful) with the noun "\textit{guł}" (flower) in Sorani Kurdish}
\label{tab_adjectives}
\end{table}

Although there are adverbs which are words in their absolute form, in most cases, adverbs are inflected forms of the absolute form of an adjective or noun in Kurdish. This process is particularly made by using the suffix "\textit{ane}" or the prefix with "\textit{be}". For instance, the two adverbs "\textit{\textbf{be}tûndî}" and "\textit{tund\textbf{ane}}" with the root "\textit{tundî}" (noun, quickness) and "\textit{tund}" (adjective, quick).

\subsection{Verbs}

In Sorani Kurdish, verbs agree with their subject in number and person and in some cases, with their object as well. For instance, in the verb "\textit{dît\textbf{im}\textbf{î}}" meaning "(I) saw you (singular)", \textit{dît}, the past stem of the verb "\textit{dîtin}" (to see), is inflected based on the enclitic pronominal logical object \textit{î} and the agent affix \textit{im}. In addition, Kurdish is a split-ergative language where subject of an intransitive verb behaves like the object of a transitive verb and differently from the agent of a transitive verb \cite{comrie1989language}. Precisely, the ergative case is marked on agents and verbs of transitive verbs in past tenses \cite{bynon1979ergative}. 

% in the initial version: "\textit{dît\textbf{im}\textbf{î}}" meaning "(I) saw him/her",

Verbs have two stems in the past and present. Although the past stem can be extracted by removing the infinitive suffix "\textit{in}", the present tense is irregularly derived from the infinitive form of the verb.

\begin{table}[h]
\centering
\scalebox{1}{
\begin{tabular}{|l|l|l|}
\hline
\textbf{Description}                                                & \textbf{Tag}  & \textbf{PoS}                                                                                                 \\ \hline \hline
\multirow{6}{*}{\parbox{4.8cm}{Connected possessive pronoun\\ (person 1-6)}} & $<$1s$>$  & \textit{im}                                                                                               \\ \cline{2-3} 
                                                           & $<$2s$>$  & \textit{it }                                                                                              \\ \cline{2-3} 
                                                           & $<$3s$>$  & \textit{î}                                                                                                \\ \cline{2-3} 
                                                           & $<$1p$>$  & \textit{man}                                                                                                 \\ \cline{2-3} 
                                                           & $<$2p$>$  & \textit{tan}                                                                                                 \\ \cline{2-3} 
                                                           & $<$3p$>$  & \textit{yan}                                                                                                 \\ \hline
\multirow{6}{*}{Copula (verbs for person 1-6)}             & $<$C1$>$   & \textit{im}                                                                                               \\ \cline{2-3} 
                                                           & $<$C2$>$   & \textit{î, ît}                                                                                            \\ \cline{2-3} 
                                                           & $<$C3$>$   & \textit{heye, hes, e  }                                                                                  \\ \cline{2-3} 
                                                           & $<$C4$>$   & \textit{în, heyn}                                                                                       \\ \cline{2-3} 
                                                           & $<$C5$>$   & \textit{in, hen }                                                                                         \\ \cline{2-3} 
                                                           & $<$C6$>$   & \textit{in, hen}                                                                                          \\ \hline
Imperative marker (verb)                                   & $<$IMP$>$  & \textit{bi, b}                                                                                               \\ \hline
Negative imperative (verb)                                 & $<$NMP$>$  & \textit{me}                                                                                                  \\ \hline
Negative marker (verb)                                     & $<$NEG$>$  & \textit{ne, na }                                                                                             \\ \hline
Subjunctive marker (verb)                                       & $<$SUB$>$  & \textit{bi}                                                                                              \\ \hline
Plural suffix (noun)                                             & $<$PL$>$   & \textit{an, gel, ha, at}                                                                                \\ \hline
Definite marker (noun)                                     & $<$DEF$>$  & \textit{eke}                                                                                       \\ \hline
Indefinite marker (noun)                                   & $<$IND$>$  & \textit{êk, yek}                                                                                      \\ \hline
Progressive marker (verb)                                  & $<$CON$>$  & \textit{de, e }                                                                                              \\ \hline
Relative adjectives      & $<$RA$>$   & \textit{î, y, în, yin, çî,}\\                                  &  &  \textit{nok, ko, û, île, yile, emenî}                                             \\ \hline
Comparative marker (adjective)                             & $<$COMP$>$ & \textit{tir}                                                                                                 \\ \hline
Superlative marker (adjective)                             & $<$SUP$>$  & \textit{tirîn}                                                                                               \\ \hline
Adverb marker (adverb)                                              & $<$AD$>$   &\textit{ane, be, an  }                                                                                       \\ \hline
&  & \textit{çe, ke, ik, ko, oke,  oł,}\\  
Diminutive form (noun) & $<$DIM$>$ & \textit{ołe, ołik, ołke, ełe, elûke, yekołe,} \\ 
&  & \textit{ îlane, île, ûlke, ûle, le, łe} \\ \hline
\end{tabular}
}
\caption{Some of the inflectional morphemes and clitics used in our finite-state analyser}
\label{cliticstable}
\end{table}

% ******************************************************************************************

\section{Implementation}
\label{implementation}

After a thorough study of our intended four grammatical categories, i.e., noun, adjectives, adverbs, and verbs, we defined their morphological features in plain language. As the result, we could extract 171 inflection rules which enabled us to obtain a deeper insight into the morphological rules for implementing transducers.

Our implementation is realized in the SFST transducer specification language which is based on regular expressions with extended functionalities such as concatenation and variable definition \cite{schmid2005programming}. The SFST compiler creates FSTs by concatenating and filtering morphemes and mapping the resulting analysis strings to the surface realizations according to phonological rules \cite{schmid2004smor}. SFST not only analyses the composing morphemes of the input word with feature decorations, but it also has a generation mode which makes the transducers reversible. This mode is particularly important for tasks relying on text generation. The following shows an example in the two modes for the word "\textit{xwardim}" (I ate):

\begin{verbatim}
analyze> xwardim
xward<verb-transitive-past-stem><past-1s>
generate> xward<verb-transitive-past-stem><past-1s>
xwardim
\end{verbatim}

The properties of base word forms are defined in a lexicon which is based on Ahmadi et al.'s work \cite{ahmadi2019lex}. In the case of nouns, adjectives and adverbs, we considered the absolute form of the word as the base form. However, this was not the case of words with derivational morphemes which was out of the scope of the current study. Regarding the verbs, two base forms are defined: the past stem and the present stem. In addition, the transitivity case of the verbs is specified as different morphological rules apply for transitive and intransitive verbs in past tenses. Figure \ref{fig_xwardin_fst} illustrates a finite-state transducer of the verb \textit{xwardin} (to eat) in the past simple tense:

\begin{figure}[h]
    \centering
    \scalebox{0.3}{    
    \includegraphics{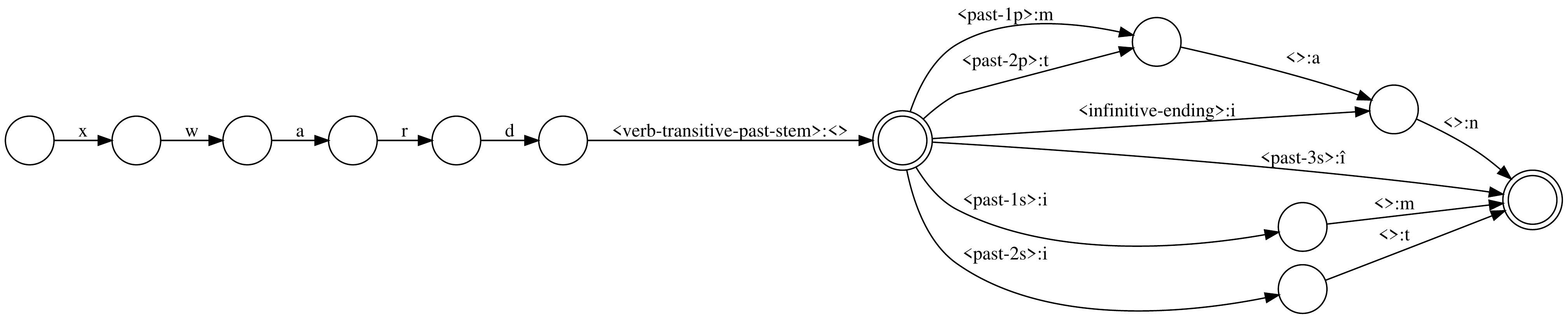}
    }
    \caption{A finite-state transducer analysing and generating simple past tense of the transitive verb \textit{xwardin} (to eat) in Sorani Kurdish}
    \label{fig_xwardin_fst}
\end{figure}

In the case of the verb, we define a basic FST composing of the verb root (in present or past) suffixed with person and number affixes. This FST enables us to reduce redundancy in defining rules as different combination of prefixes such as \textit{bi}, \textit{ne} and \textit{de} respectively as progressive, negative and imperative prefixes, would be feasible. Having said that, the ergative cases are not covered in this initial study. Table \ref{cliticstable} provides some of the inflectional morphemes which are used as variable in our transducers.

Kurdish is written using different scripts in which Persian-Arabic and Latin are the two most widely used. The challenges in the processing of Persian-Arabic texts~\cite{ahmadi2019wergor}, particularly the missing "\textit{i}" vowel, also known as \textit{Bizroke} among Kurdish scholars, is the main reason that in the current work we focused on Latin-based script. 

\section{Conclusion and future work}
\label{conclusion}

In this paper, we reported our progress in utilizing FST for the morphological analysis of Kurdish. We presented the four main categories of the Kurdish morphology, namely verbs, nouns, adjectives and adverbs. This research equips Kurdish NLP and CL with an advanced tool and resource which improves the quality of the Kurdish language processing tasks.  

In this preliminary study, our developed tool can analyse and generate rather simple morphological forms. Therefore, as future work, we are aiming to cover more linguistic processes such as ergativity, and other grammatical categories in Sorani Kurdish. Moreover, analysing morphological derivation of Kurdish should be a priority, particuarly in the case of verbs which are mostly formed based on derivational morphemes. Another limitation of the current study is the absence of syntactical features which may modify morphological forms in a sentence. 

In our implementation, we only used the Latin script of Kurdish as it is the least ambiguous one in comparison to the Persian-Arabic script \cite{ahmadi2019wergor}. We believe that the current study will pave the way for evaluating machine transliteration systems in future as well. In addition, we believe that the computation analysis of the other dialects of Kurdish will pave the way for further progress in the field.

Our tool and resource will be publicly available for non-commercial use under the CC BY-NC-SA 4.0 licence upon the acceptance of the paper.

\bibliographystyle{ACM-Reference-Format}
\bibliography{bibliography}

\end{document}